\documentclass{llncs}
\usepackage{graphicx}
\usepackage{url}
\usepackage{subfig}

\title{An Empirical Comparison of Syllabuses for Curriculum Learning}
\author{Mark Collier \and Joeran Beel}
\institute{Trinity College Dublin \\
\email{\{mcollier,joeran.beel\}@tcd.ie} }

\begin{document}
\maketitle

\begin{abstract}
Syllabuses for curriculum learning have been developed on an ad-hoc, per task basis and little is known about the relative performance of different syllabuses. We identify a number of syllabuses used in the literature. We compare the identified syllabuses based on their effect on the speed of learning and generalization ability of a LSTM network on three sequential learning tasks. We find that the choice of syllabus has limited effect on the generalization ability of a trained network. In terms of speed of learning our results demonstrate that the best syllabus is task dependent but that a recently proposed automated curriculum learning approach - \textit{Prediction Gain}, performs very competitively against all identified hand-crafted syllabuses. The best performing hand-crafted syllabus which we term \textit{Look Back and Forward} combines a syllabus which steps through tasks in the order of their difficulty with a uniform distribution over all tasks. Our experimental results provide an empirical basis for the choice of syllabus on a new problem that could benefit from curriculum learning. Additionally, insights derived from our results shed light on how to successfully design new syllabuses.
\end{abstract}

\section{Introduction}

Curriculum learning is an approach to training neural networks inspired by human learning. Humans learn skills such as maths and languages by tackling increasingly difficult tasks. It has long been proposed \cite{RN40,mitchell1980need} that neural networks could benefit from a similar approach to learning. Curriculum learning involves presenting examples in some order during training such as to aid learning. On some tasks, curriculum learning has been shown to be necessary for learning \cite{RN12,RN44,RN59} and to improve learning speed \cite{RN41,RN43} and generalization ability \cite{RN39} on other tasks.

For any particular application of curriculum learning the two key components of this approach to training neural networks are the division of the training data into tasks of varying difficulty and the definition of a syllabus. We follow \cite{RN41} in defining a \textit{syllabus} to be a ``time-varying sequence of distributions over tasks". Where a \textit{task} is defined to be a subset of all training examples. At any instance during training a batch called an \textit{example} is drawn from some task and presented to the network. For example in machine translation we could split the training data of source and target language pairs into \textit{tasks} defined by the length of the source sentence. We could then define a \textit{syllabus} in which the network was first trained on short sentences and gradually moving onto longer, potentially more difficult sentences.

In this work we focus on the effect of syllabus design on curriculum learning. Syllabuses fall in two main categories. \textit{Hand-crafted syllabuses} \cite{RN43,RN44,RN12,RN39} define an ordering of tasks and a level of success on each task to be attained before moving onto the next point in the syllabus. The level of success on a task is typically evaluated periodically on a validation set.

\textit{Automated syllabuses} \cite{RN41,RN45} attempt to solve the problem of having to hand-design a syllabus and choosing the level success that determines progression from task to task required when using a hand-crafted syllabus. Automated syllabuses only require the engineer to define a set of tasks (potentially ordered) and examples are chosen by some automatic mechanism e.g. in proportion with the error rate on that task.

The choice of syllabus has a significant effect on the efficacy of curriculum learning for a particular problem \cite{RN41,RN43}. Syllabuses have been primarily developed on an ad-hoc basis for individual problems and only limited empirical comparison of different syllabuses has been conducted \cite{RN41,RN43}. Additionally explanations for why curriculum learning works have mostly been grounded in learning theory \cite{RN40,RN46,RN39} rather than empirical results.

We consider curriculum learning in two settings \cite{RN41}. In the \textit{multi-task setting}, after training we are concerned with the performance of the network on all tasks. Whereas in the \textit{target-task setting}, after training we are concerned only with the performance of the network on a single ``target'' task.

In this work we evaluate how the choice of syllabus effects the learning speed and generalization ability of neural networks in the multi-task and target-task settings. We train LSTM \cite{RN18} networks on three synthetic sequence learning tasks \cite{RN11}. Sequence learning tasks are naturally suited to curriculum learning as it is often assumed that the difficulty of the task increases with the length of the input sequence \cite{RN43,RN59,cirik2016visualizing,spitkovsky2010baby,RN12}. LSTMs have achieved state of the art performance in many sequence learning tasks \cite{RN36,RN16,RN37}. It is common to use LSTMs trained on synthetic tasks for experimental results in curriculum learning \cite{RN43,RN41,cirik2016visualizing,spitkovsky2010baby,RN46}.

Our results provide an extensive empirical comparison of the effect of curriculum learning syllabuses on speed of learning and generalization ability. Our results reveal insights on why curriculum learning outperforms standard training approaches and the key components of successful syllabuses. Thus our work provides evidence for the choice of syllabus in new applications of curriculum learning and provides the basis for future work in syllabus design.

\section{Related Work}

Bengio, et al. \cite{RN39} examined curriculum learning under hand-crafted syllabuses. They found that a Perceptron trained on a toy task with a syllabus which steps through tasks in their order of difficulty learns faster than a Perceptron trained on randomly sampled examples. Their work demonstrated the potential utility of curriculum learning, but the division between curriculum learning and transfer learning was unclear in their experiments and only a basic hand-crafted syllabus was considered.

Zaremba and Sutskever \cite{RN43} showed that curriculum learning was necessary for a LSTM network to learn to ``execute'' simple Python programs. The authors provide an empirical comparison of two hand-crafted syllabuses. Their problem had two dimensions of difficulty the length of the program and the level of nesting. They defined a syllabus which they call the \textit{Combined Strategy} that combines the basic approach of stepping through the tasks one at a time with a uniform distribution over all tasks. The Combined Strategy syllabus lead to faster learning and better final performance on the target task than baseline strategies.

The authors hypothesize that by including examples from tasks more difficult than the current task, the LSTM network has an incentive to use only a portion of its memory to solve the current task in the syllabus. The authors argue that syllabuses which do not provide tasks from future points on the syllabus incur a significant time penalty at each point of progression as the network must undergo substantial retraining of the network weights to adapt to the new task. In a separate work, the same authors proposed a slightly more complex syllabus with the same goal of sampling tasks based on a syllabus with ``non-negligible mass over the hardest difficulty levels'' which enabled reinforcement learning of a discrete Neural Turing Machine \cite{RN55}, where a network trained directly on the target task failed to learn.

Zaremba and Sutskever only considered syllabuses with a linear progression through the tasks in order of difficulty. Yet it has been argued that linear progression through the tasks in a syllabus can result in quadratic training time in the length of the sequences for a given problem \cite{RN59}. Following a simple syllabus of drawing tasks from a uniform distribution over all tasks up to the current task and exponentially increasing the level of difficulty of the problem at each progression point it is possible to scale a variant of the Neural Turing Machine \cite{RN11} to sequences of length 4,000 \cite{RN59}.

Graves, et al. \cite{RN41} propose an automated syllabus by treating the selection of the task from which to draw the next example from as a stochastic adversarial multi-armed bandit problem \cite{RN54}. Their paper examines six definitions of a reward function to promote learning. The reward functions split into two classes, those driven by the improvement in the loss of the network as a result of being trained on an example from a particular task and those driven by an increase in network complexity motivated by the Minimum Description Length principle \cite{rissanen1986stochastic,grunwald2007minimum}. The authors show that the best reward function from each class results in non-uniform syllabuses that on some tasks converge to an optimal solution twice as fast as a baseline uniform sampling syllabus. Across all tasks however the authors find that the uniform sampling syllabus provides a strong baseline. Other automated syllabuses have been used successfully \cite{RN45,matiisen2017teacher}.

Curriculum learning is not widely applied in practical machine learning and has not always been successfully applied in research settings either  \cite{avramova2015curriculum}. ``Catastrophic forgetting" of previously learned tasks may be responsible for poor generalization performance after training \cite{mccloskey1989catastrophic}. Developing syllabuses robust to catastrophic forgetting may be critical to achieving wider application of curriculum learning.

\section{Methodology}

\subsection{Syllabuses}

We considered six syllabuses that captured the main features of those identified in the literature. The six syllabuses include three hand-crafted, one automated and two benchmark syllabuses. Below, we follow the notation that there are \textit{T} tasks on the syllabus that are ordered in difficulty from 1 to T and that the learner's current task is denoted C. A full distribution over tasks for each syllabus is given in table \ref{syllabus_dist}.

\begin{table}
\centering
\caption{Distribution over tasks for syllabuses.}\label{syllabus_dist}
\begin{tabular}{|l|l|l|l|l|}
\hline
Syllabus &  C & Uniform(1, max\{1, C-1\}) & Uniform(1, T) & T\\
\hline
Naive & 100\% & - & - & - \\
Look Back & 90\% & 10\% & - & - \\
Look Back and Forward & 80\% & - & 20\% & - \\
None & - & - & - & 100\%\\
Uniform & - & - & 100\% & - \\
Prediction Gain & - & - & - & - \\
\hline
\end{tabular}
\end{table}

Under the Naive Strategy proposed by by Zaremba and Sutskever \cite{RN43} examples at each timestep are sampled solely from the current point C on the syllabus. Once the learner reaches the defined level of success for progression on the current task as measured on a validation set, the learner moves onto the next task on the syllabus and all examples are now drawn from that task. We call this syllabus \textit{Naive}.

It was observed that while the Naive syllabus increased the rate of learning on certain tasks, the learner rapidly forgot previously learned tasks \cite{RN45}. This is undesirable in the multi-task setting and it was proposed \cite{RN12} that drawing some examples from previous tasks on the syllabus may prevent this catastrophic forgetting. We define a syllabus which we call \textit{Look Back} in which a fixed percentage of examples are drawn from a Uniform distribution over all previous tasks on the syllabus and the remainder are drawn from the current task on the syllabus as per the Naive syllabus. In practice for our experiments we chose to draw 10\% of tasks from previous points on the syllabus, with the remaining 90\% coming from the current task.

While the Look Back syllabus addresses the issue of catastrophic forgetting, it was further hypothesized that by only drawing examples from the current and past tasks in the syllabus considerable retraining of the network weights would be required as the learner moved forward through the syllabus \cite{RN43}. A syllabus which we call \textit{Look Back and Forward} addresses this issue. Look Back and Forward corresponds to the syllabus which Zaremba and Sutskever call the Combined Strategy \cite{RN43}. With the Look Back and Forward syllabus a fixed percentage of examples are drawn from a Uniform distribution over all tasks on the syllabus with the remaining examples drawn from the current task. Thus, when on the early tasks in the syllabus almost all examples drawn from the uniform distribution will be drawn from future tasks on the syllabus. Once the learner approaches the target task almost all such examples will be drawn from previously learned tasks. In this way the Look Back and Forward syllabus seeks to address both the issue of catastrophic forgetting and the potential retraining required when the learner is not given examples from upcoming tasks on the syllabus. In our experiments, we drew 20\% of examples from the Uniform distribution over all tasks, with the remaining 80\% coming from the current task on the syllabus. We note that for the Look Back and Look Back and Forward syllabuses, we chose the percentage splits through experimentation.

We adopt the best performing automated syllabus consistent with maximum likelihood training from recent work on automated curriculum learning \cite{RN41}, which the authors call \textit{Prediction Gain}. The authors follow their general approach of selecting the next task based on the Exp3.S algorithm \cite{RN54} for stochastic adversarial multi-armed bandits. The reward function for Prediction Gain is defined to be the scaled reduction in loss $L$ on the same example $x$ after training on that example, i.e. the reward to the bandit is defined to be $L(x, \theta) - L(x, \theta')$ rescaled to the $[-1,1]$ range by a simple procedure \cite{RN41}, where $\theta$ and $\theta'$ are the weights parameterizing the network before and after training on $x$ respectively.

\textit{None} is our benchmark syllabus in the target-task setting, for which we draw all examples from the target task in the syllabus.

As our benchmark syllabus in the multi-task setting we draw examples from a Uniform distribution over all tasks. Unsurprisingly, we call this syllabus \textit{Uniform}. This syllabus can also be seen as a simple syllabus in the target-task setting and as mentioned above has been found to be a strong benchmark in this setting \cite{RN41}.

In practice for the above hand-crafted syllabuses we, as other authors have found \cite{RN59}, that learning was slow if the learner progressed through the syllabus one task at a time. To alleviate this, for each of these syllabuses after meeting the defined success metric on a task instead of moving onto the next task in the syllabus we followed an exponential strategy of doubling the current point on the syllabus along one dimension of difficulty on the problem. For the Repeat Copy problem (described below) which has two dimensions of difficulty, we alternated which dimension of difficulty to double.

\subsection{Benchmark Problems}

As noted above curriculum learning has primarily been applied to sequence learning problems where the sequence length is typically used to divide the data into tasks ordered by difficulty. We follow this approach by adopting three synthetic sequence learning problems that have been shown to be difficult for LSTMs to solve \cite{RN11}.

\textbf{Copy} - for the Copy problem, the network is fed a sequence of random bit vectors followed by an end of sequence marker. The network must then output the input sequence. This requires the network to store the input sequence and then read it back from memory. In our experiments we trained on input sequences up to 32 in length with 8-dimensional random bit vectors.

\textbf{Repeat Copy} - similarly to the Copy problem, with Repeat Copy the network is fed an input sequence of random bit vectors. Unlike the Copy problem, this is followed by a scalar that indicates how many times the network should repeat the input sequence in its output sequence. In our experiments we trained on input sequences up to 13 in length with maximum number of repeats set to 13, again with 8-dimensional random bit vectors. This means the target task for Repeat Copy required an output sequence of length 169.

\textbf{Associative Recall} - Associative Recall is also a sequence learning problem, with sequences consisting of random bit vectors. In this case the inputs are divided into items, with each item consisting of 3 x 6-dimensional vectors. After being fed a sequence of items and an end of sequence marker, the network is then fed a query item which is an item from the input sequence. The correct output is the next item in the input sequence after the query item. We trained on sequences of up to 12 items, thus our target task contained input sequences of 36 6-dimensional random bit vectors followed by the query item.

\subsection{Experiments}

For all of the above problems we ran experiments\footnote{Source code to replicate our experiments can be found here: \url{https://github.com/MarkPKCollier/CurriculumLearningFYP}} to measure the effect of syllabus choice on speed of learning in both the target-task and multi-task setting and the generalization ability of the trained network. For each problem, syllabus pair we ran training from 10 different random initializations.

In order to measure the speed of learning we measured the performance during training every 200 steps on two held-out validation sets, one for the target-task and one for the multi-task setting. The validation set for the target-task setting consisted solely of examples from the target task on that problem. Whereas, the validation set for the multi-task setting consisted of examples uniformly distributed over all tasks in the syllabus. The number of examples in the target-task and multi-task validation sets were 512 and 1024 respectively for all experiments.

To test the effect of syllabus choice on the generalization ability of our networks, for each problem we created test sets of 384 examples which gradually increased the difficulty of the problem. For the Copy problem which was trained with a target task of sequences of length 32, the test set comprised on sequences of length 40.

For Repeat Copy we wished to test the generalization ability of the trained networks along both dimensions of difficulty of the problem - sequence length and number of repeats. We created two test sets, one comprised of sequences of length 16 with the number of repeats fixed to 13. The other test set comprised of sequences of length 13 with the number of repeats set to 16. Our test set for Associative Recall consisted of sequences of 16 items, having been trained with a target task of 12 items. We measured the performance of each network on the test set when the network's error on the validation set was lowest.

In all our experiments we used a stacked 3 x 256 units LSTM network, with a cross-entropy loss function. For all networks and syllabuses we used the Adam optimizer \cite{RN15} with an initial learning rate of 0.01. We used a batch size of 32 for all experiments, with each batch containing only examples from a single task. We defined the maximum error achieved on a task before progression to the next point on the syllabus to be 2 bit errors per sequence for the Copy problem and 1 bit error per sequence on the Repeat Copy and Associative Recall problems.

\section{Results}

\subsection{Copy}

\begin{figure}%
    \centering
    \subfloat[target-task setting]{{\includegraphics[width=0.45\textwidth]{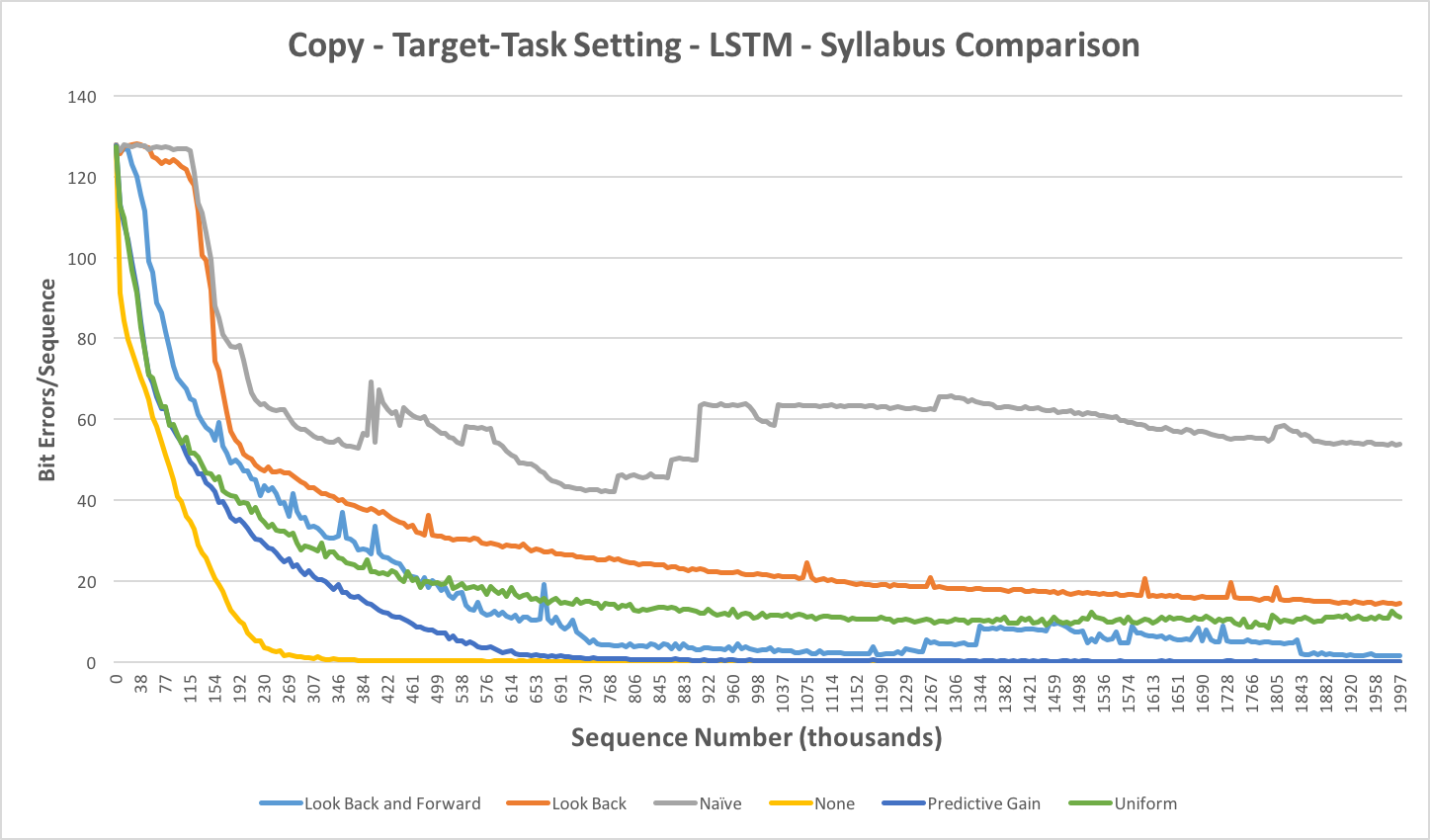} }}%
    \qquad
    \subfloat[multi-task setting]{{\includegraphics[width=0.45\textwidth]{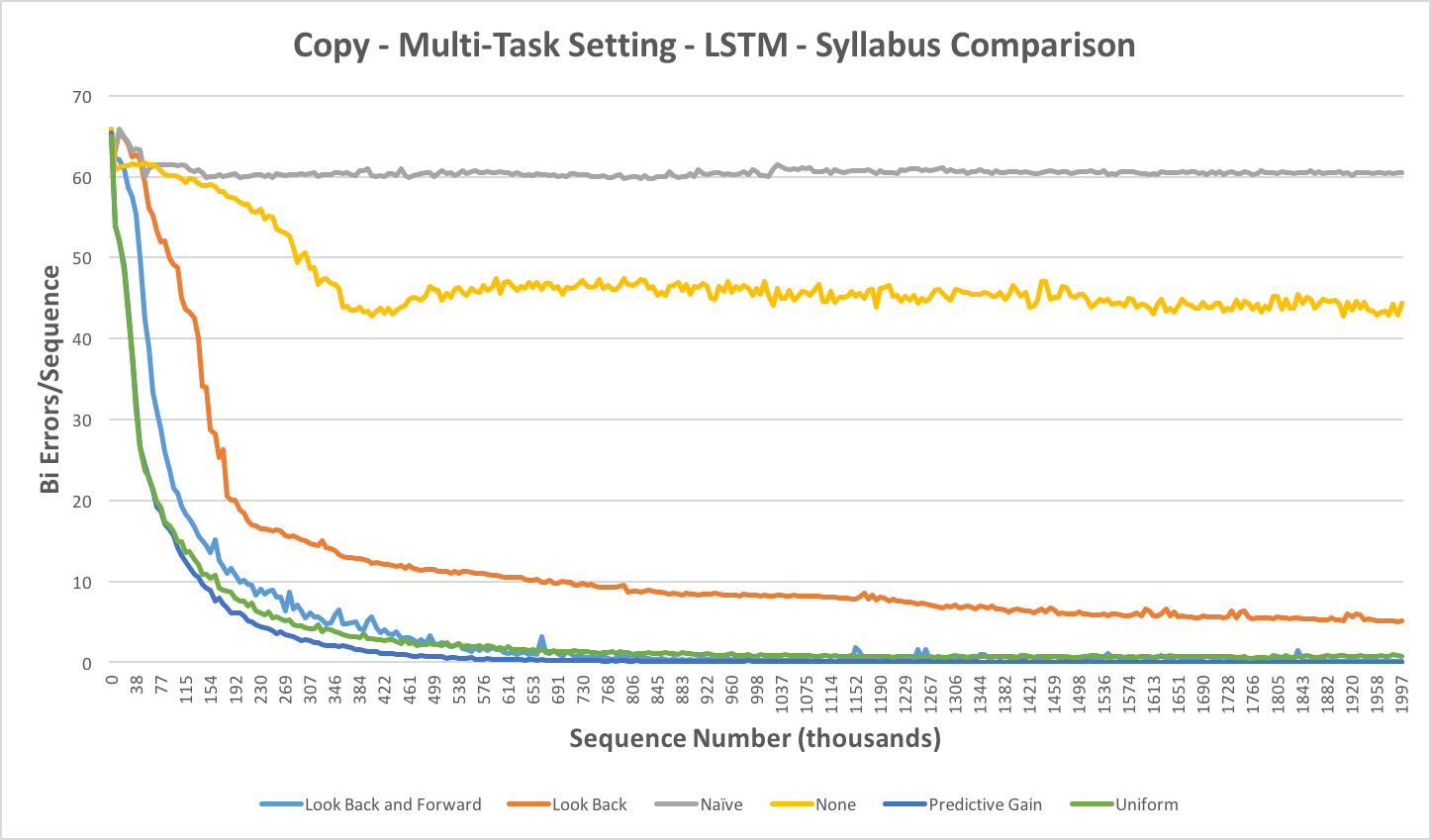} }}%
    \caption{Median Copy learning curves}%
    \label{fig:copy_learning_curves}%
\end{figure}

Figure \ref{fig:copy_learning_curves} shows the median learning curves over the 10 training runs per syllabus for the target-task and multi-task setting. Training directly on the target task converges to near zero error ~2.2 times faster than the next fastest syllabus - Prediction Gain. This demonstrates that in the target-task setting a syllabus is not required to solve the task. Prediction Gain adapts most rapidly to this situation and learns to focus on tasks close to the target task. Only the Look Back and Forward syllabus manages to reach near zero error in the target-task setting in the time provided. All the hand-crafted syllabuses must reach the same level of success on a task to progress so the performance of Look Back and Forward relative to the other hand-crafted syllabuses shows that sampling from tasks ahead of the current task in the syllabus allows much faster progression through the tasks.

As expected, in the multi-task setting training directly on the target task performs very poorly when evaluated on all tasks. Despite making some progress towards the target task, the Naive syllabus shows almost no improvement on random performance in the multi-task setting. This demonstrates that the network rapidly forgets previously learned tasks if not shown further examples of them. Prediction Gain, Uniform and Look Back and Forward all converge to near zero error in the multi-task setting at similar rates, fig. \ref{fig:copy_learning_curves}.

\begin{figure}
\centering
\includegraphics[width=0.45\textwidth]{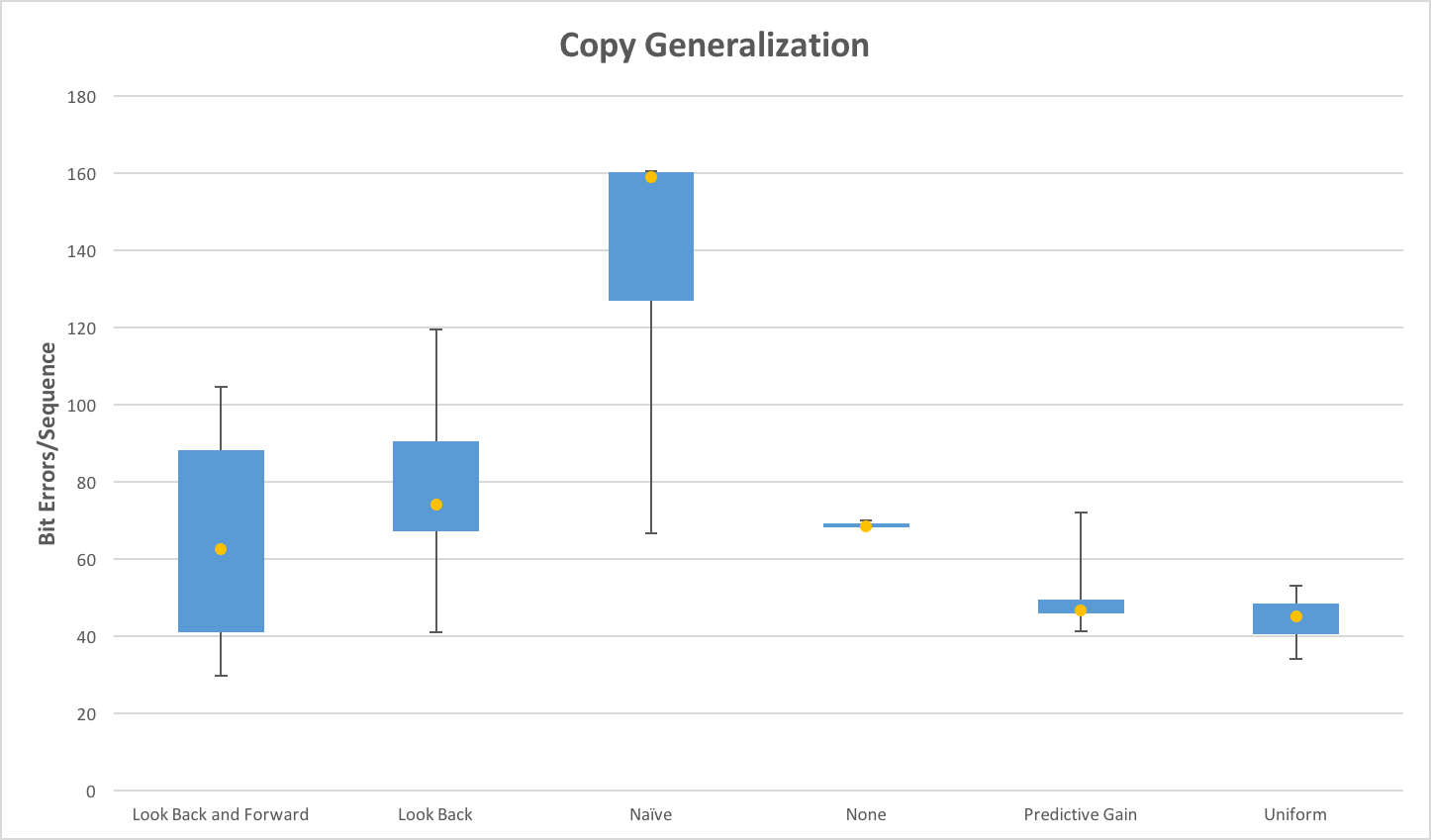}
\caption{Copy - Generalization to sequences of length 40.  Each box plot gives the median (yellow dot), upper and lower quartiles (blue box) and min and max error (black lines) on the 10 training runs.}
\label{fig:copy_generalization}
\end{figure}

Figure \ref{fig:copy_generalization} shows the range of error achieved by the 10 trained networks for each syllabus when the networks are asked to generalize to sequences of length 40 on the Copy problem. Despite the networks of three syllabuses converging to near zero error on the target task with sequences of length 32 none of the networks succeed in generalizing to sequences of length 40. On sequences of length 40, the Prediction Gain and the Uniform syllabus demonstrate similar performance and have approximately 1.35-1.59 times lower median error than the Look Back and Forward, None and Look Back Syllabuses. There is substantial overlap in the range of generalization error for all syllabuses, so we cannot say that any one syllabus clearly outperforms the others in terms of improving generalization on the Copy problem.

\subsection{Repeat Copy}

\begin{figure}%
    \centering
    \subfloat[target-task setting]{{\includegraphics[width=0.45\textwidth]{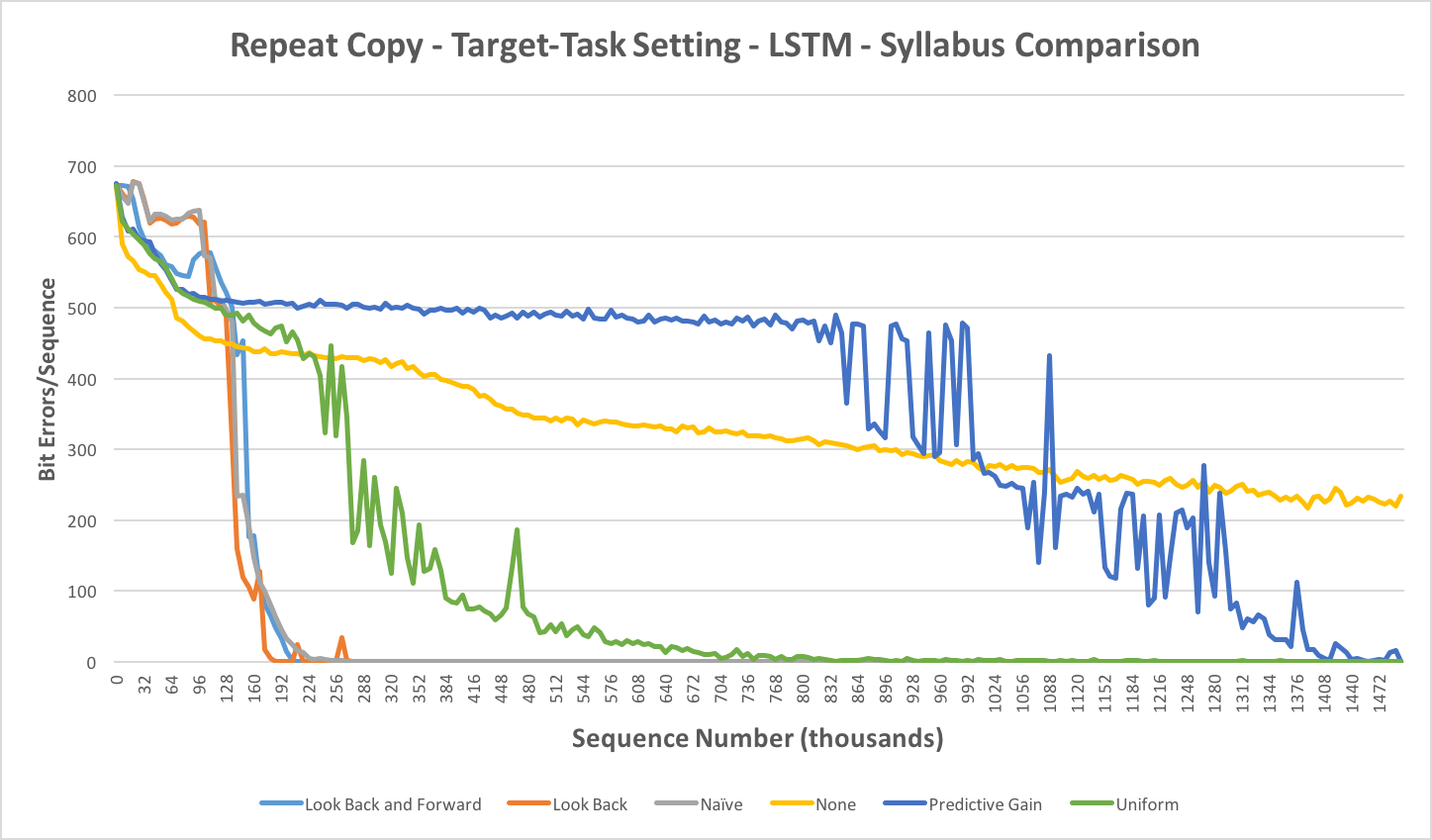} }}%
    \qquad
    \subfloat[multi-task setting]{{\includegraphics[width=0.45\textwidth]{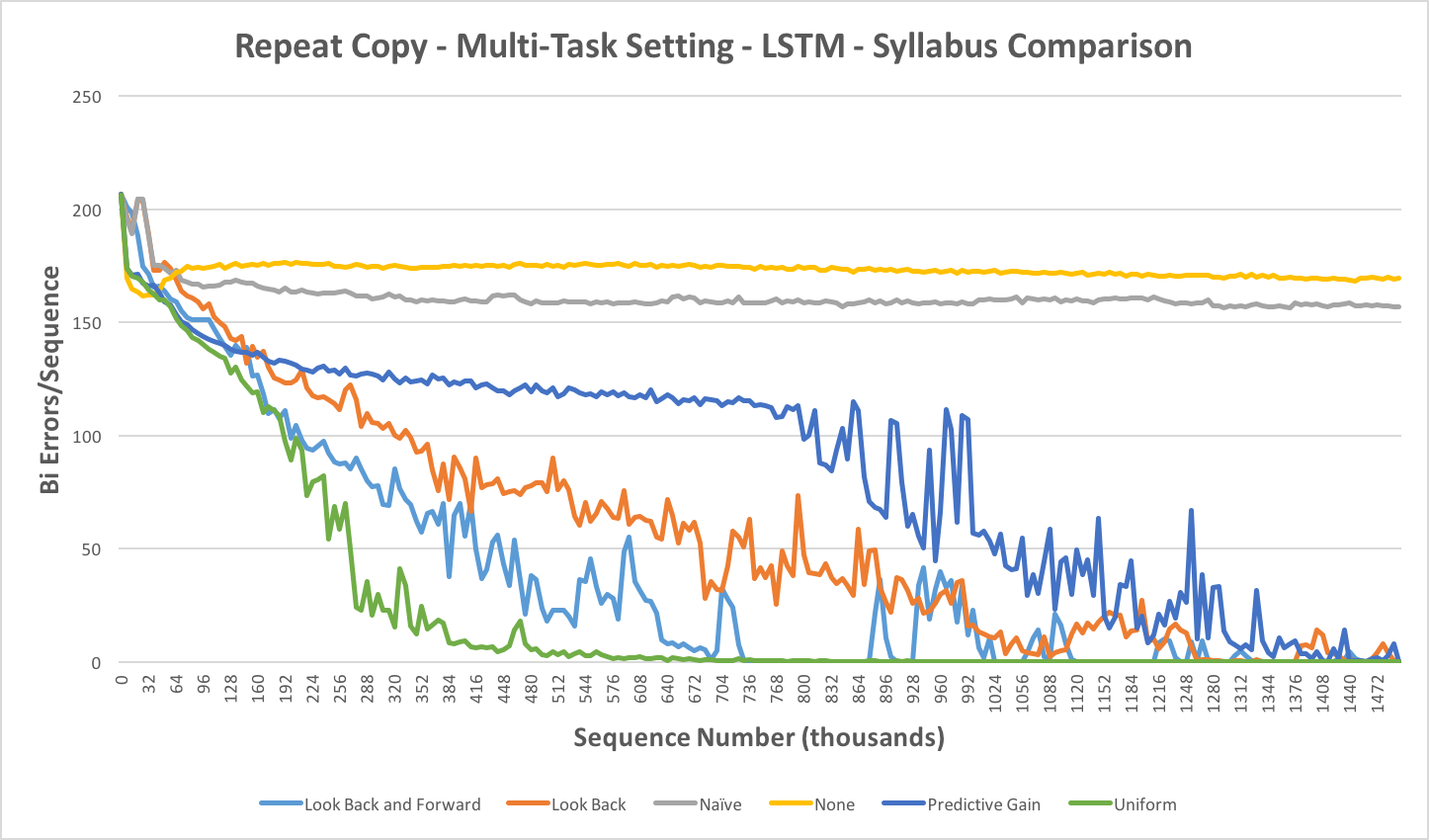} }}%
    \caption{Median Repeat Copy learning curves}%
    \label{fig:repeat_copy_learning_curves}%
\end{figure}

Figure \ref{fig:repeat_copy_learning_curves} shows the median learning curves for each syllabus on the Repeat Copy problem for the target-task and multi-task setting. Unlike for the Copy problem In the target-task setting for the Repeat Copy problem, the network fails to learn a solution in the time provided by training directly on the target task, fig. \ref{fig:repeat_copy_learning_curves}. Interestingly the three hand-crafted syllabuses converge to near zero error at approximately the same time and ~3.5 times faster than the Uniform syllabus which is the next fastest. This is a clear win for the hand-crafted syllabuses over the benchmark and automated syllabuses.

The Uniform syllabus converges to near zero error twice as fast as Prediction Gain. It is unclear why Prediction Gain has slower convergence although we posit a potential explanation - that scaling the rewards by the length of the input sequence as per the specification \cite{RN41} may bias the bandit towards tasks with high repeats as such tasks incur no penalty for their added difficulty. This highlights that despite the automated nature of Prediction Gain's syllabus generation it still relies on several tunable parameters.

Despite the Uniform syllabus' slow convergence on the target task, in the multi-task setting training on the same distribution as the test distribution is beneficial, as would be expected in non curriculum learning settings, fig. \ref{fig:repeat_copy_learning_curves}. The Uniform syllabus reaches near zero error ~1.35 times faster than Look Back and Forward, the next fastest syllabus.

\begin{figure}%
    \centering
    \subfloat[Generalization to input sequences of length 16, with the number of repeats fixed to 13]{{\includegraphics[width=0.45\textwidth]{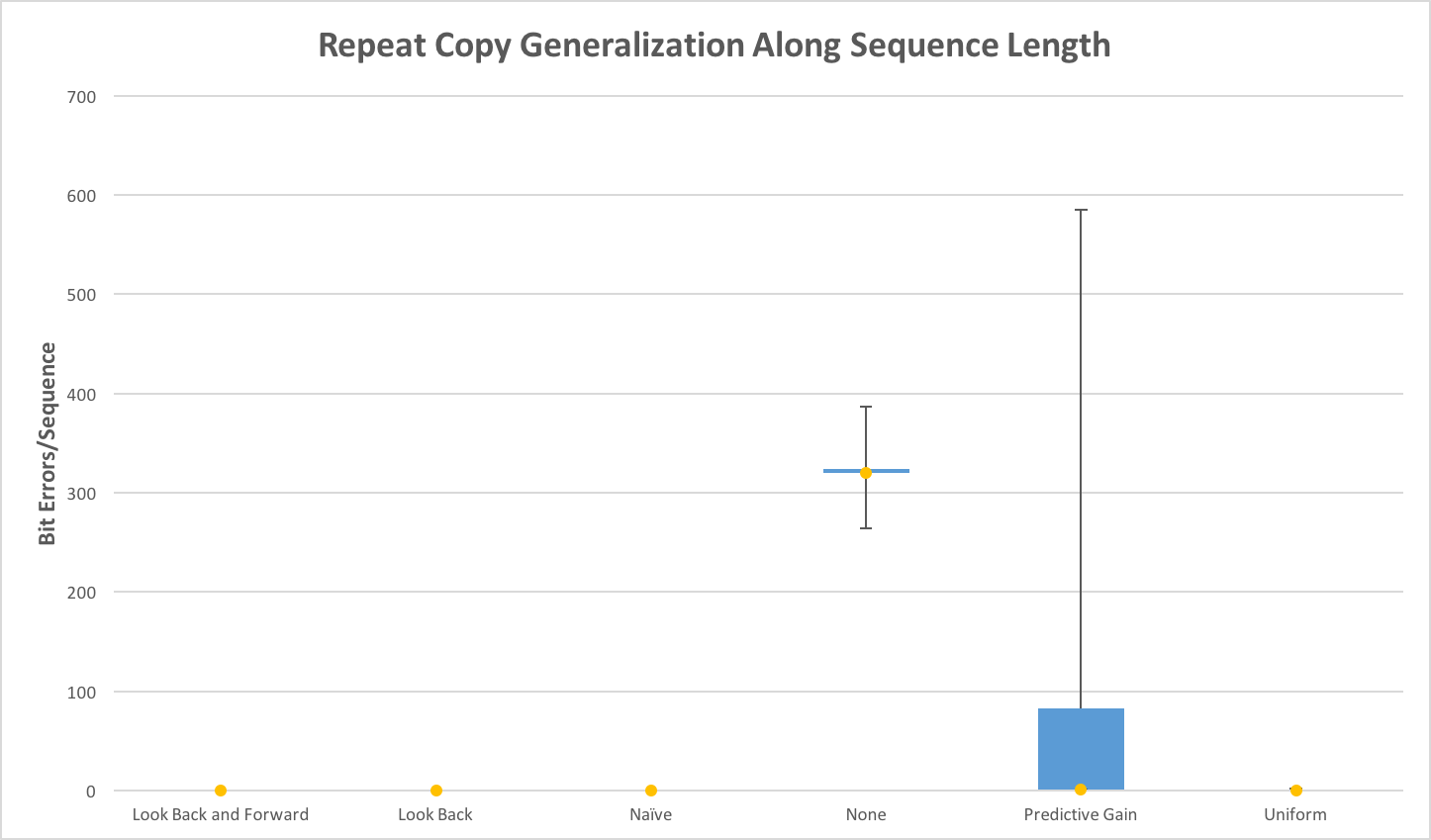} }}%
    \qquad
    \subfloat[Generalization to 16 repeats, with the input sequence fixed to length 13]{{\includegraphics[width=0.45\textwidth]{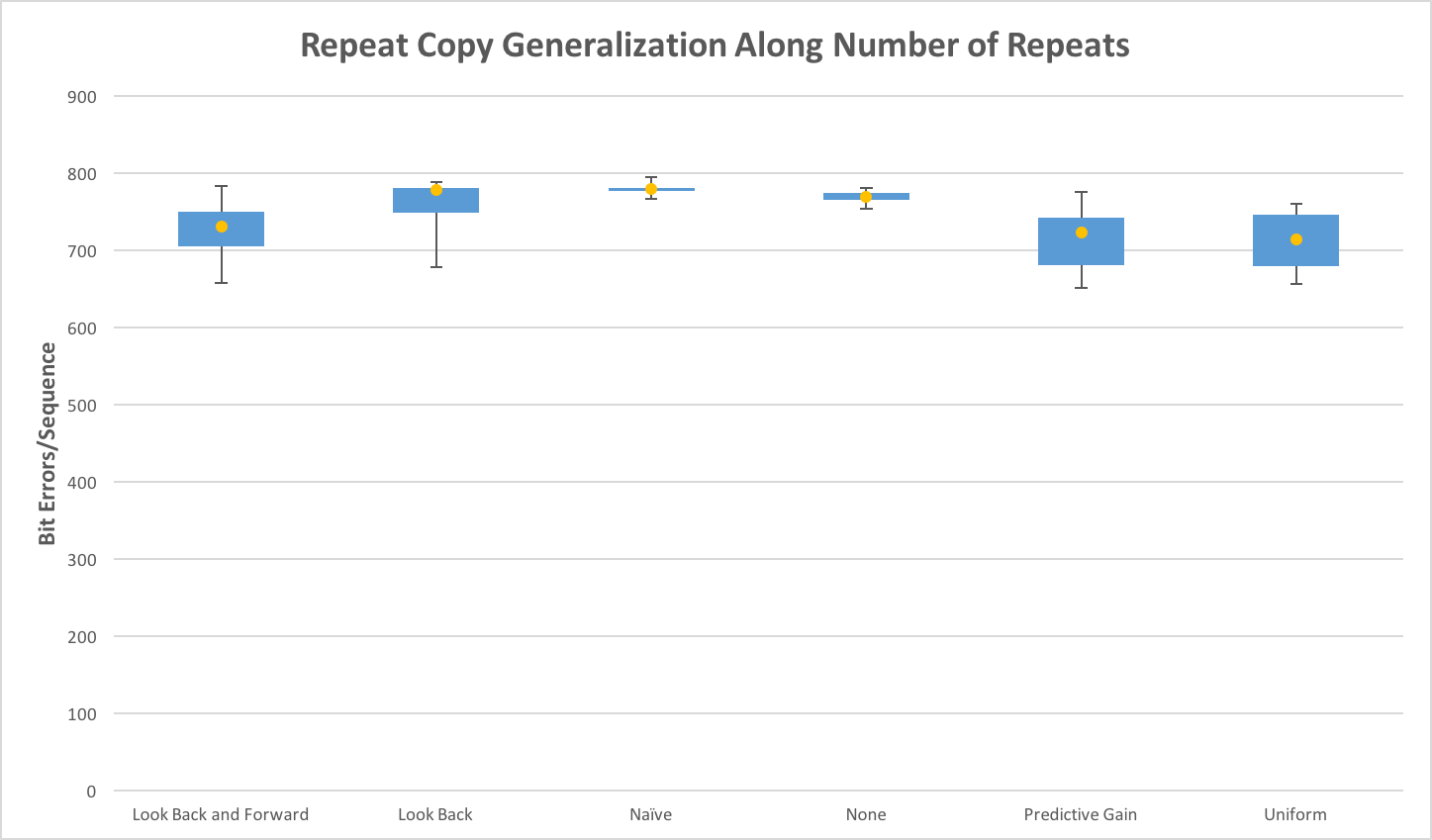} }}%
    \caption{Repeat Copy - Generalization performance.}%
    \label{fig:repeat_copy_generalization}%
\end{figure}

All the syllabuses but Prediction Gain and None consistently generalize with near zero error to sequences of length 16 on Repeat Copy, fig. \ref{fig:repeat_copy_generalization}. Prediction Gain's wide generalization error range is explained by the fact that training with Prediction Gain on the Repeat Copy problem is unstable and 4 out of the 10 training runs failed to converge to near zero error. Whereas when we attempt to increase the number of repeats, all the syllabuses fail to generalize successfully, fig. \ref{fig:repeat_copy_generalization}, this is consistent with previous results \cite{RN41}.

\subsection{Associative Recall}

\begin{figure}%
    \centering
    \subfloat[target-task setting]{{\includegraphics[width=0.45\textwidth]{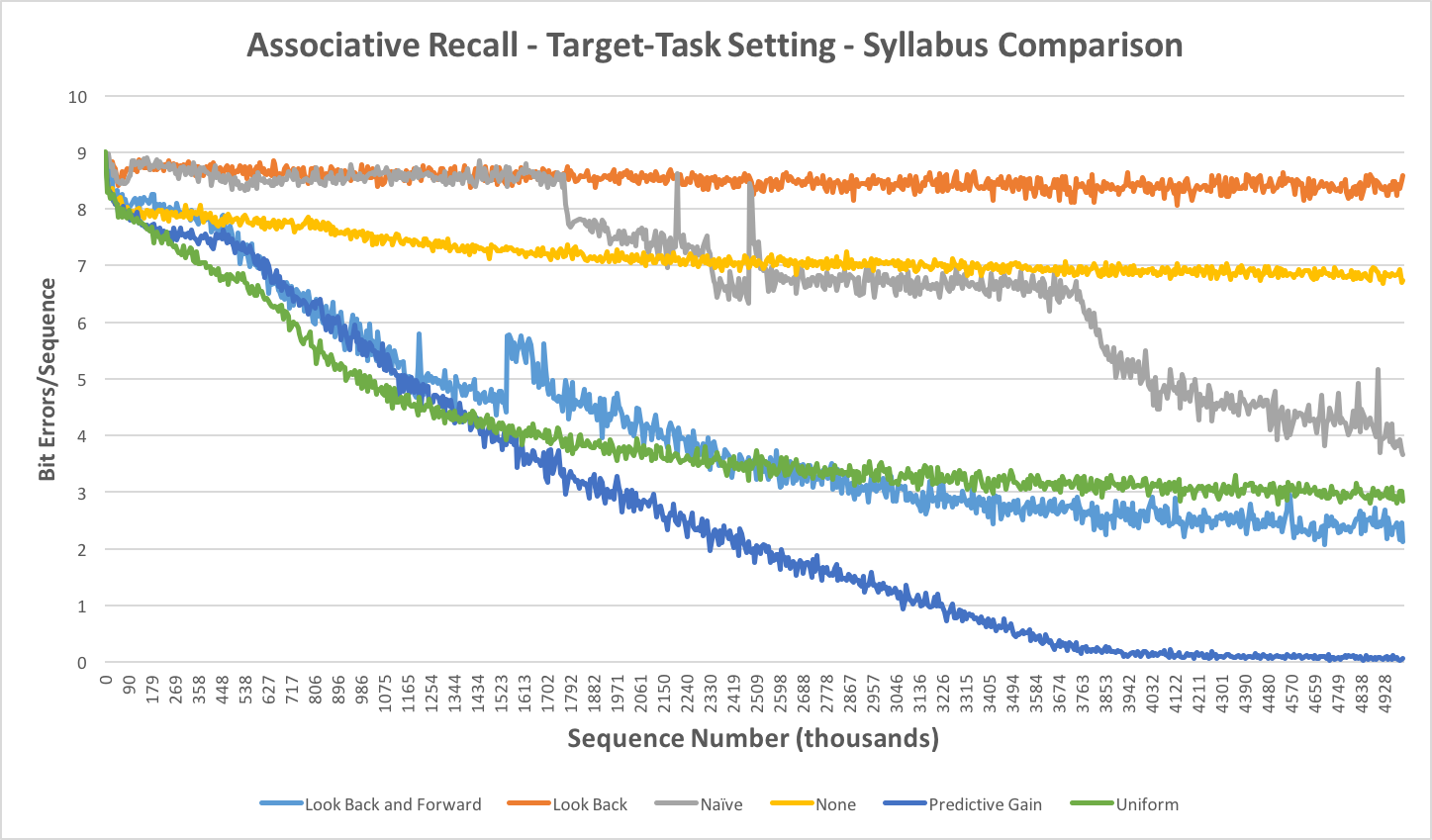} }}%
    \qquad
    \subfloat[multi-task setting]{{\includegraphics[width=0.45\textwidth]{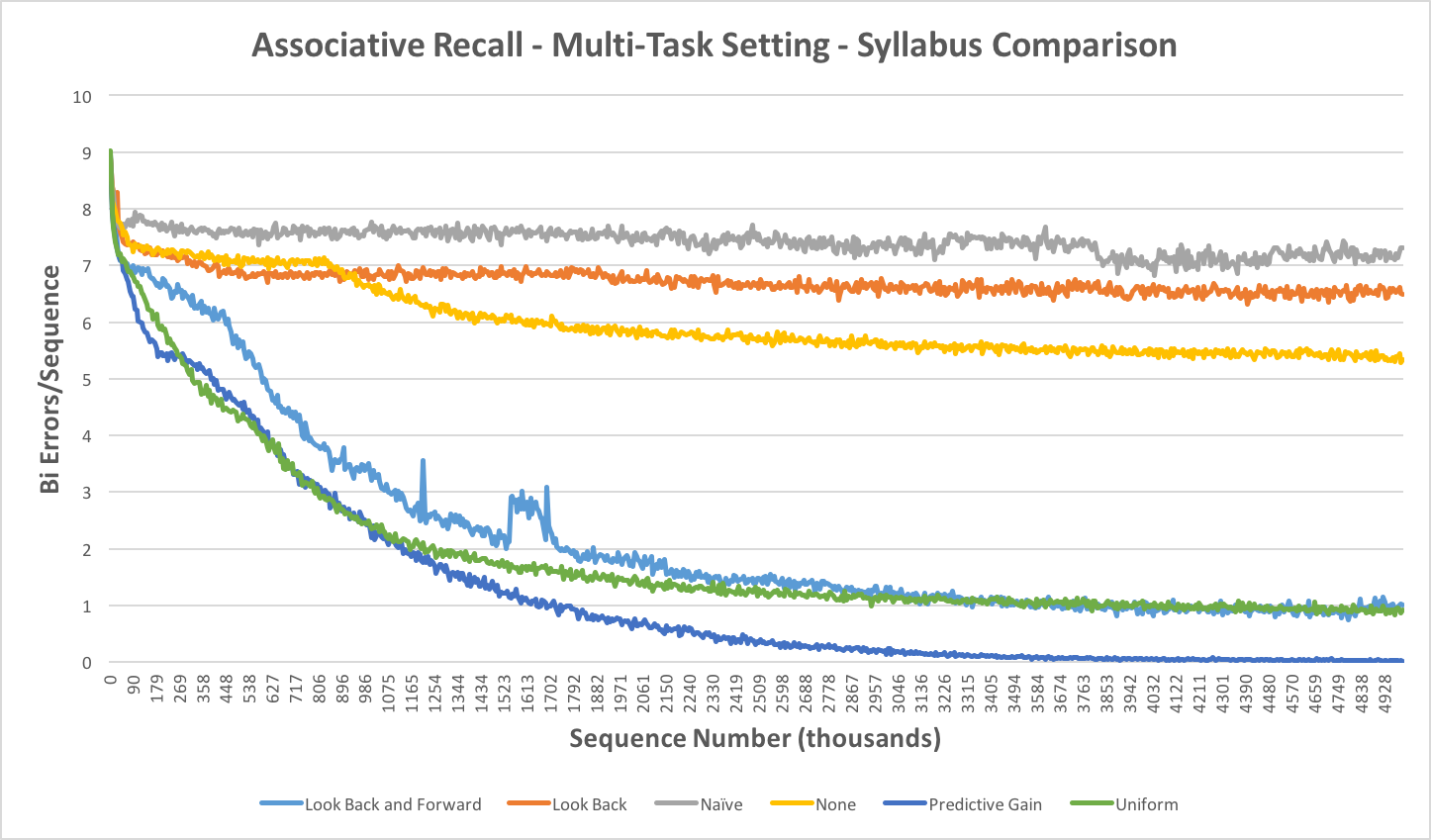} }}%
    \caption{Median Associative Recall learning curves}%
    \label{fig:associative_recall_learning_curves}%
\end{figure}

Figure \ref{fig:associative_recall_learning_curves} shows the median learning curves for each syllabus on Associative Recall for the target-task and multi-task setting. Only Prediction Gain converges to a near zero error solution to the Associative Recall problem in either setting. The Uniform and Look Back and Forward syllabuses perform similarly in both settings and make some progress towards a low error solution. The Naive syllabus makes some progress towards learning the target-task and perhaps given enough time would learn a solution to the task. The Look Back and None syllabuses make very limited progress from random initialization in either setting.

\begin{figure}
\centering
\includegraphics[width=0.45\textwidth]{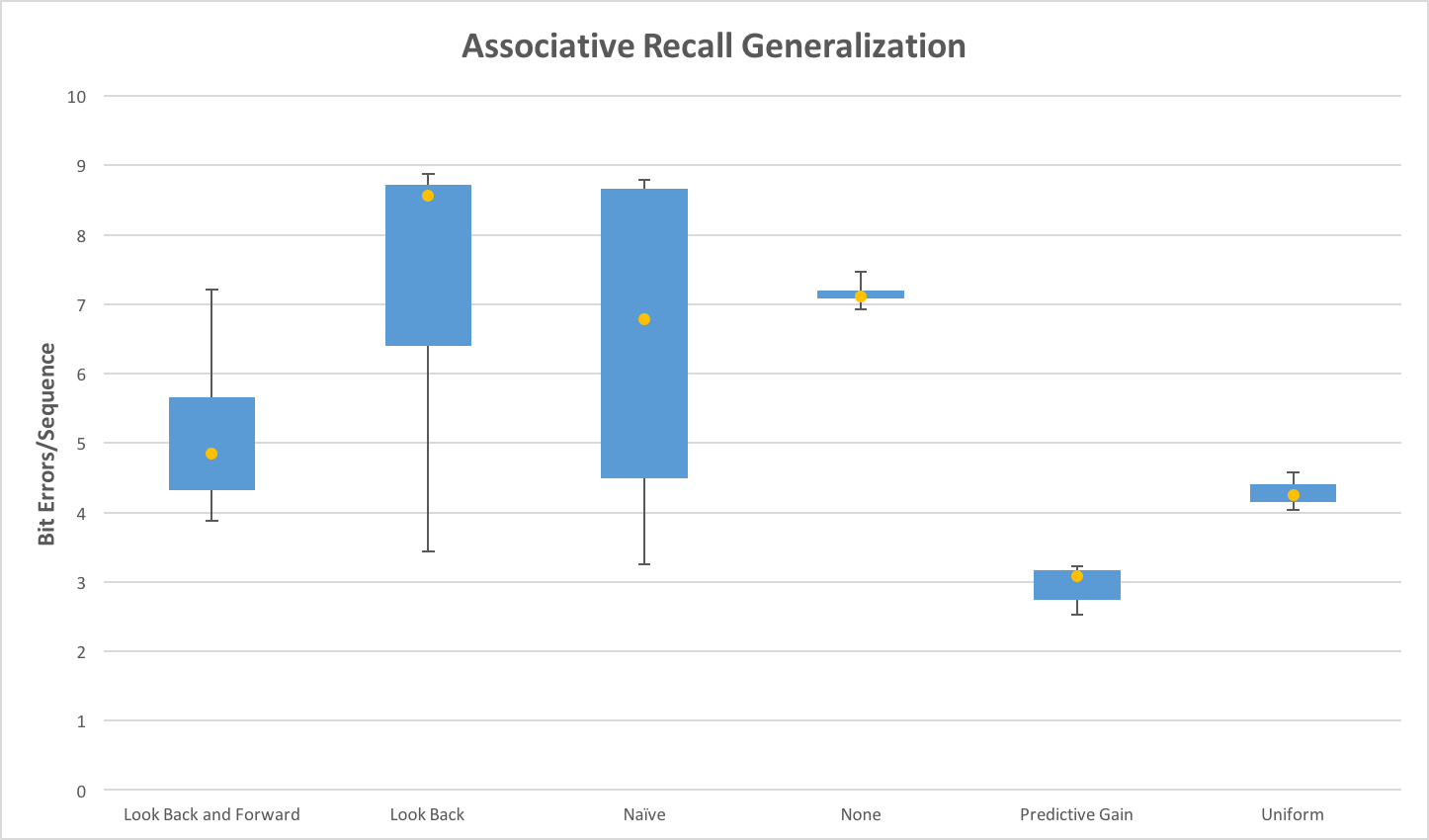}
\caption{Associative Recall - Generalization to sequences with 16 items.}
\label{fig:associative_recall_generalization}
\end{figure}

Despite converging to a near zero error solution in both the target-task and multi-task setting, figure \ref{fig:associative_recall_generalization} shows that Prediction Gain fails to generalize with similar error to sequences of 16 items. As expected, the other syllabuses which do not reach near zero error on the target task also exhibit similarly high generalization error.

\section{Conclusion}



The above experiments demonstrate that no single syllabuses in our comparison has an advantage in all problems (Copy, Repeat Copy, Associative Recall) or settings (target-task, multi-task). Over the three problems, the Look Back and Forward syllabus learned consistently faster than the other two hand-crafted syllabuses; Look Back and Naive. Prediction Gain was fastest of the non-benchmark syllabuses to converge to near zero error solutions on the Copy and Associative Recall problems but the opposite was true on Repeat Copy (when training with Prediction Gain was unstable). We found that the Uniform syllabus provided a strong baseline in the target-task setting, which agrees with previous results \cite{RN41}. In the multi-task setting uniformly sampling from all tasks leads to the fastest learning in two of the three problems. No syllabus consistently improved generalization on any of the above problems.

Including examples from previously learned tasks is vital to prevent catastrophic forgetting of these tasks. For example, the Naive syllabus which makes progress towards the target-task on each benchmark problem, is among the two slowest learners in the multi-task setting on all problems. This demonstrates that the network rapidly forgets previously learned tasks when it is no longer presented with them.

Including examples from future tasks on the syllabus provided substantial increases in speed of learning. In particular, the Look Back and Forward syllabus converged to a lower error solution in both settings on all problems than the Look Back syllabus which does not include examples from future tasks on the syllabus but is otherwise identical.

Despite being compared to stronger syllabuses than in the original paper, Prediction Gain performed competitively. We conclude that automated approaches to syllabus design in curriculum learning may be a fruitful future area of development but that further work is required on how to set the hyperparameters governing the existing automated approaches.

Our empirical results provide the basis for the choice of syllabus in new applications of curriculum learning. In the multi-task setting we recommend using the non curriculum learning syllabus of uniformly sampling from all tasks. Other syllabuses may provide marginal gains on some problems in the multi-task setting, but this is not reliable across all problems and requires additional hyperparameter tuning. We recommend that when applying curriculum learning to problems in the target-task setting, practitioners use either Prediction Gain or the Look Back and Forward syllabus.

\subsubsection{Acknowledgements}

This publication emanated from research conducted with the financial support of Science Foundation Ireland (SFI) under Grant Number 13/RC/2106.

\bibliographystyle{splncs03}

\bibliography{aics-sample}

\end{document}